# Structured Pattern Pruning Using Regularization

Dongjun Park, Geung-Hee Lee

**Abstract**— Iterative Magnitude Pruning (IMP) is a network pruning method that repeats the process of removing weights with the least magnitudes and retraining the model. When visualizing the weight matrices of language models pruned by IMP, previous research has shown that a structured pattern emerges, wherein the resulting surviving weights tend to prominently cluster in a select few rows and columns of the matrix. Though the need for further research in utilizing these structured patterns for potential performance gains has previously been indicated, it has yet to be thoroughly studied. We propose SPUR (Structured Pattern pruning Using Regularization), a novel pruning mechanism that preemptively induces structured patterns in compression by adding a regularization term to the objective function in the IMP. Our results show that SPUR can significantly preserve model performance under high sparsity settings regardless of the language or the task. Our contributions are as follows: (i) We propose SPUR, a network pruning mechanism that improves upon IMP regardless of the language or the task. (ii) We are the first to empirically verify the efficacy of "structured patterns" observed previously in pruning research. (iii) SPUR is a resource-efficient mechanism in that it does not require significant additional computations.

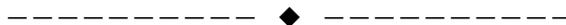

## 1 INTRODUCTION

Transfer learning using transformer models has shown state-of-the-art performance in a broad range of tasks in natural language processing such as, and not limited to, NLI (Natural Language Inference) and NLG (Natural Language Generation).

However, these transformer-based models require dozens of times more parameters than the existing RNN and CNN-based models, making them prohibitively expensive for deployment to devices with limited memory, CPU, and storage space such as smartphones and IOT infrastructure. In order to overcome these challenges, various models of network pruning methods have been thoroughly researched.

Iterative Magnitude Pruning (IMP) is a network pruning method that repeats the process of pruning the weights with the smallest values and retraining the model [1]. IMP is a widely-used industry standard because it not only delivers high pruning efficiency but it is parallelly applicable with other compression mechanisms such as quantization and knowledge transfer.

As recent research has shown, IMP has an interesting property that a certain structured pattern emerges as a result of the compression. Visualizations of the inner weights of NLP models that have been pruned by IMP show that the surviving weights end up prominently clustered in a select few sets of rows and columns of the weight matrices, and this has been shown to occur without any particular penalties on specific rows and columns to induce such behavior [2]. It has been inferred that there exists a relationship between these structured patterns and the efficacy of IMP when used in NLP models. Clearly, this warrants further research in exploiting these structured patterns in pruning deep learning models. However there has not been related research on this topic thus far [3].

Besides these developments, recent research has proposed a Movement Pruning technique [4], which has shown to achieve up to 95% of original performance while removing 97% of model weights by considering the magnitude of the gradient instead of the weights themselves. Though such results are promising, two limitations generally exist: 1) Since experiments are only conducted in English NLP tasks, there is a need for additional verification of value when applied to multilingual tasks. 2) Such mechanisms require additional gradient matrix computations and training iterations, leading to significant additional burden regarding initialization settings and computing resources.

Consequently, this study proposes SPUR (Structured Pattern pruning Using Regularization), a network pruning mechanism that preemptively induces structured patterns by adding a regularization term to the loss function of IMP. Our research empirically studies whether SPUR can significantly retain the performance of transformer models [5], [6] when applied to NLI and QA (Question Answering) tasks in English, Korean, and French settings [7]–[12].

This study contributes the following: (i) We propose SPUR, a mechanism that improves the performance of IMP regardless of the language or the task at hand. (ii) We are the first to empirically verify the efficacy of structured patterns in network pruning observed in previous research [2]. (iii) The method we propose is highly resource-efficient in that it requires no significant additional opti-

- *Dongjun Park is with Shinhancard, Seoul, Korea, KR 04551. E-mail: broaddeep@gmail.com.*
- *Geung-Hee Lee is with the Department of Statistics and Data Science, Korea National Open University, Seoul, Korea, KR 03087. E-mail: geunghee@knou.ac.kr.*

mizations and considerations of initialization settings, leading to relatively faster optimization speeds and lower memory costs.

The paper is organized as follows. In Section 2 we review related literature. We describe the proposed method in Section 3. In Section 4 we evaluate the performance of SPUR with additional experiments. We conclude in Section 5.

## 2 RELATED WORK

The Iterative Magnitude Pruning (IMP) was initially proposed in the context of pruning image processing models [1]. Though specific implementations may vary in details, the IMP generally proceeds with the following procedure [13]. First, it determines the target removal rate at the current point, 50% for example. Second, it sets a reference value such that the amount of weights removed will be 50%. Third, the weights below the reference value are removed. Fourth, the model is retrained to recover the decreased model performance as a result of the compression. Fifth, the process returns to Step 1 with an increased removal rate, of 60% for example. Finally, the procedure is terminated when the target removal rate has been achieved.

The reference value can be defined in the context of the full combined set of weights of the entire network or ranked separately within individual weight matrices. The former is denoted as a global pruning and the latter as the local pruning, and this study focuses on the latter approach. In other words, in the local pruning paradigm for this study, the proportion of weights removed is uniform across individual weight matrices that make up the network.

In the context of transformer models, the first attempted iterative magnitude pruning is the RPP (Reweighted Proximal Pruning) [2]. RPP is a standard IMP procedure in that it considers the magnitude of the weights as the removal criterion, and it utilizes the concept of $\ell_1$ regularization [14] to remove unnecessary weight values. As typical $\ell_1$ regularization induces parts of the weights to converge to 0, this has the consequence that the weights with greater magnitudes experience greater proportional penalties compared to lower ones. Considering the general pruning principle of maximally preserving the most important weights of the network, given a set of weights with high magnitudes in a particular step, there is a need to reduce the penalty applied upon them in the following steps to reduce their consideration. For this purpose, instead of a standard $\ell_1$ regularization term applied on the entire set of weights $w_i$, the objective function weighted with importance values $\alpha_i$ was established as in (1).

$$\underset{\mathbf{w}}{\text{minimize}} \; f_0(\mathbf{w}) + \gamma \sum_i \alpha_i |w_i| \qquad (1)$$

$$\alpha_i^t = \frac{1}{|w_i^t|^{(t)} + \epsilon}$$

In this case, $f_0$ is a typical loss function, $\alpha_i^t$ and $w_i^t$ rep-resent $\alpha_i$ and $w_i$ at time $t$, respectively, and $\epsilon$ is a very small real number. That is, if the weight $w_i$ of the previous step is large, the size of the denominator also increases and $\alpha_i$ becomes small, thereby establishing an inversely proportional relationship.

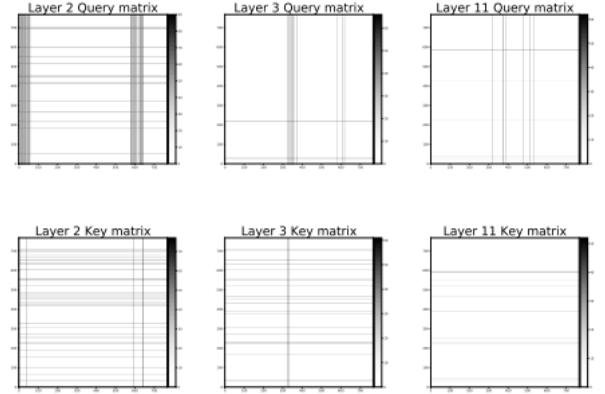

Fig. 1. Structured patterns observed in Reweighted Proximal Pruning[2]

As shown in Fig. 1, visualizations of the inner weights of models pruned by RPP have often shown that the surviving weights tend to prominently cluster in particular sets of rows and columns of the matrix. Following research reviews [3] have argued that researching the effectiveness of using these structured patterns for model compression is necessary.

Recently, another study that has been attracting attention in the field of iterative pruning is Movement Pruning [4]. As a pruning technique based on the magnitude of the change in weights, since it does not consider the absolute magnitude of the weights themselves, it is difficult to regard it as another IMP. Particularly in high-sparsity cases where more than 85% of the encoder inner weights are removed, MVP consistently outperforms standard IMP methods. Despite its performance gains, it has a critical limitation that it requires the addition and training of score matrices of the same size as the full set of the weight matrices; this adds an often prohibitively compounded amount of complications regarding initialization settings that vary between the problem context (e.g. the particular task and the language) and heavily increased computing resources.

In conclusion, a review of a pruning technique that utilizes the structured patterns that arise as a result of IMP methods is necessary. Additionally, a pruning method that competently retains the model performance without the addition of initializing and optimizing separate score matrices is desired. This study aims to achieve this purpose by adding a regularization term that can preemptively induce the structured patterns that emerge in network pruning.

## 3 PROPOSED METHOD

This study proposes the following objective functions and removal criteria to preemptively induce the aforementioned structured patterns.

Firstly, the training objective aims to optimize the loss function $L$ to a minimum. This loss function $L$ not only contains the cross-entropy loss $L_{ce}$ that represent the difference between model predictions and the ground truth, but it also induces a regularization loss $L_R$ term related to the deviance of the weights from their desired values. Specifically, the importance of $L_R$ is adjusted to the size of $\lambda$.

$$L = L_{ce} + \lambda L_R \quad (2)$$

### 3.1 Regularization Loss

This study intends to set $L_R$ so that weights end up distributed in a structure that is advantageous for network pruning. Each of the weight matrices constituting BERT $W \in R^{r \times c}$ is a two-dimensional matrix with r rows and c columns. BERT's encoder consists of a total of 12 layers, and each layer contains 6 dense matrices. Of these, four correspond to multihead self-attention modules, and two correspond to feed-forward networks (FFNs).

We evaluate our methods on BERT not only by applying it to all six matrices, but also by applying it on specific subsets of them. This was done to evaluate whether significant performance differences would result in varying the range of weight matrices that are targeted by the pruning mechanism. Specifically, aside from the full $BERT$ model, we also apply it on $BERT_{Q+K}$ (applied only on query ($Q$) and key ($K$) matrices of BERT's self-attention modules), $BERT_Q$ (only the $Q$ matrix), and $BERT_K$ (only the $K$ matrix).

$L_R$ is a value computed first by summing the deviance $D(W)$ for all target matrices $W$ then standardizing by $N(BERT)$, the total number of target matrices. Therefore, as the number of matrices with large deviance increases, the value of the overall loss function $L$ increases, and the model is trained to lower the deviance of the matrix.

$$L_R = \frac{1}{N(BERT)} \sum_{W \in BERT} D(W) \quad (3)$$

### 3.2 Deviance

Deviance $D(W)$ is computed by first calculating the deviance of the absolute values of the weights from their expected values $E(|W_{i,j}|)$. Then these values are standardized by dividing them with the square root of their expected values. We then end up with the final desired balanced sum of squares value by dividing the sum with $N(W) = r \times c$.

$$D(W) = \frac{1}{N(W)} \sum_{i=1}^{r} \sum_{j=1}^{c} \left\{ \frac{|W_{i,j}| - E(|W_{i,j}|)}{\sqrt{E(|W_{i,j}|)}} \right\}^2 \quad (4)$$

This deviance $D(W)$ can also be viewed as applying an $\ell_2$ penalty on the sum of the standardized deviance of each weight value. Since the performance may be affected by the particular choices of this penalty, we experiment with variant forms of $D(W)$. $D_{L1S}(W)$ applies an $\ell_1$ penalty on the standardized deviance of the weights and $D_{L1}(W)$ applies it on the deviance without standardization. On the other hand, $D_{L2}(W)$ imposes an $\ell_2$ penalty on non-standardized deviations.

$$D_{L1S}(W) = \frac{1}{N(W)} \sum_{i=1}^{r} \sum_{j=1}^{c} \left| \frac{|W_{i,j}| - E(|W_{i,j}|)}{\sqrt{E(|W_{i,j}|)}} \right| \quad (5)$$

$$D_{L1}(W) = \frac{1}{N(W)} \sum_{i=1}^{r} \sum_{j=1}^{c} \left| |W_{i,j}| - E(|W_{i,j}|) \right| \quad (6)$$

$$D_{L2}(W) = \frac{1}{N(W)} \sum_{i=1}^{r} \sum_{j=1}^{c} \left\{ |W_{i,j}| - E(|W_{i,j}|) \right\}^2 \quad (7)$$

### 3.3 Expected Value

The expected value $E(|W_{i,j}|)$ is a value whose denominator is the sum of absolute values of the weight matrix and the numerator is the product of the sum of the rows and columns of absolute values. As a result, it is expected that weights located in the same row and column will be induced to have similar values. It has the advantage of being simple to implement and capable of intuitive interpretation.

$$E(|W_{i,j}|) = \frac{(\sum_{m=1}^{c} |W_{i,m}|)(\sum_{n=1}^{r} |W_{n,j}|)}{\sum_{m=1}^{c} \sum_{n=1}^{r} |W_{n,m}|} = \frac{(row\ sum)(col\ sum)}{grant\ total\ sum} \quad (8)$$

### 3.4 Pruning Criteria and Scheduling Function

In addition to these training goals, the removal criteria follow the local pruning approach as indicated above. That is, in accordance to a reference sparsity $v\%$, the elements $|W_{i,j}|$ in the top $v\%$ within its *own* weight matrix survive and the rest are removed. The sparsity $v^{(t)}$ at time t is calculated and adjusted by the cubic scheduling function consisting of the current step $t$, the initial sparsity $v_{initial}$, the final sparsity $v_{final}$, and the total number of steps $T$. This is the same as the implementation of the Movement Pruning [4]. In addition, the magnitude of $\lambda$ also varies according to the cubic scheduling function as the steps progress.

$$v^{(t)} = v_{final} + (v_{initial} - v_{final})\left(1 - \frac{t - t_i}{N \Delta t}\right)^3 \quad (9)$$

In this study, in the same way as other state-of-the-art pruning techniques [4], [15], even if a specific weight has been removed due to its small size at a particular step, it can be reselected in a future step.

## 4 EXPERIMENTS

### 4.1 Experimental Setup

In this study, we experiment with two prominent tasks in NLP: Natural Language Inference (NLI) and Question Answering (QA). Firstly, NLI is a task widely known for involving semantic containment and exclusion between texts. Given a pair of sentences, the objective is to classify whether their relationship is that of entailment, neutrality, or contradiction. A model attempting to solve this task must have the ability to identify referential relationships, tenses, and grammatical ambiguities found in text.

Firstly, a variety of representative NLI datasets are available for many individual languages. For this study we experiment with the MNLI (Multi NLI) dataset [7] for English and the KorNLI (Korean NLI) dataset [9], as these are the most representative large-scale datasets in their respective languages. For experiments with French, we simply select the French components from the XNLI (Cross-Lingual NLI) dataset [8].

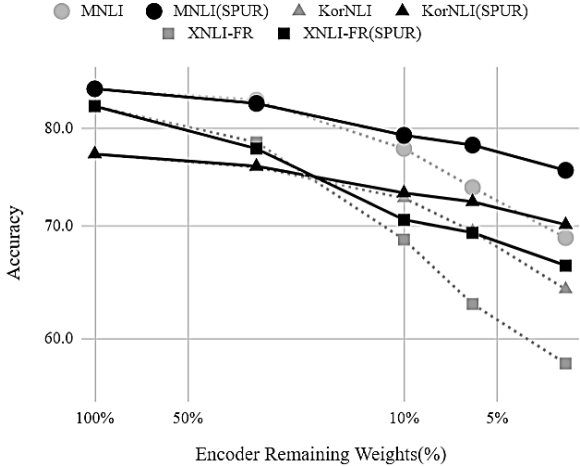
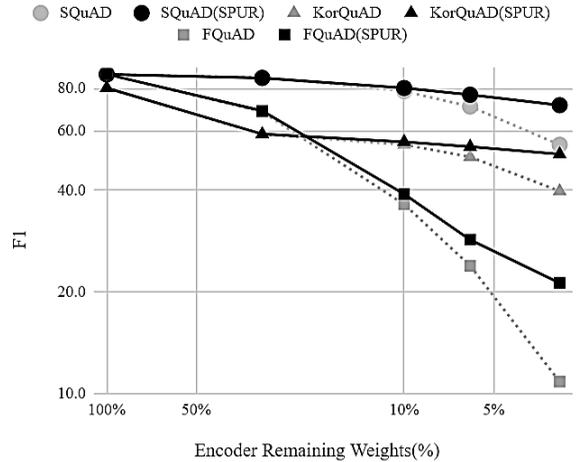

Fig. 2. Performance comparison on NLI task. Dotted line corresponds to only using IMP and solid line to additionally applying SPUR.

Fig. 3. Performance comparison on QA task. Dotted line corresponds to only using IMP and solid line to additionally applying SPUR.

Secondly, Question Answering (QA) tasks are a type of machine reading comprehension evaluation that requires reading long pieces of text and answering questions regarding it. For this task, the SQuAD (Stanford Question Answering Dataset) is the most representative dataset curated with over 100,000 question-answer pairs grounded in Wikipedia articles. For our experiments, we use the SQuAD dataset [10] for English, KorQuAD (Korean Question Answering Dataset) [12] for Korean and FQuAD (French Question Answering Dataset) [11] for French. We specifically use the 1.1 version of SQuAD.

In regard to pretrained transformer models, we use BERT [6] for English, KoBERT [16] for Korean, and CamemBERT [17] for French. These three models are structurally identical in relation to their encoder layer compositions, with the significant difference being that they were constrained to their respective languages in their semi-supervised pre-training procedures. Another difference can be found in the size of the embedding matrices, since the number of possible tokens vary between languages. However, the encoder layers where the vast majority of model weights can be found are identical in structure and size across these three models.

### 4.2 Implementation

The basic code for performing the experiments was implemented by partially modifying the publicly available code for Movement Pruning [4]. Optimization configurations such as the sparsity scheduling function, hyperparameters such as the learning rate, and the target weights subjected to pruning are kept as close as possible to the source as the following. For NLI, batch sizes were all set to 32 and the learning rate was set to 5e-5 for Korean and 3e-5 for the other languages. Initial and final warm-up steps were both set as 1 epoch and the models were all trained for 6 epochs. For Question Answering, the batch size and learning rate were set uniformly across the models as 16 and 3e-5, respectively. The English and Korean models were trained for 10 epochs and the French model was trained for 14 epochs. Initial and final warm up steps were set at 1 and 2 epochs respectively for both Korean and English, and for French it was 3 and 4 epochs, respectively. We increased the number of epochs for the French model since it experienced performance gains due to its smaller QA dataset size.

The BERT model used in the experiments consists of an embedding matrix and an encoder; regardless of the language, the encoder consists of 12 layers, where each layer consists of 6 dense layers. Among these 6 layers, four correspond to the Multi-Head Self Attention module and the rest to the FFN (Feed-Forward Network) module. The embedding weights were frozen during training, and either the entirety of the encoder layers or subsets of them were subject to pruning (for the purpose of ablation analysis).

SPUR includes a regularization loss $L_R$ in addition to the standard loss function. Specifically, since the importance of $L_R$ is adjusted by the size of $\lambda$, the selection of this hyperparameter warrants deeper consideration. For this reason, we experimented with $\lambda$ values of 10 and 100 for encoder sparsities of 3%, 6%, 10%, and 30%, and selected the $\lambda$ values yielding the most optimal performance after pre-experimenting with heuristically selecting $\lambda$ values

### 4.3 Experimental Results

Fig. 2 and Fig. 3 show that pruning the network with SPUR leads to a greater preservation of original model performance than that of standard IMP methods. The dotted line shows the case where only IMP was applied, and the solid line shows the case where SPUR is also applied. In particular, the performance gap between SPUR

TABLE 1
NLI PERFORMANCE COMPARISON BY SPUR

| Encoder Remaining Weights(%) | Method | Natural Language Inference (Accuracy) | | |
|---|---|---|---|---|
| | | English (MNLI) | Korean (KorNLI) | French (XNLI-FR) |
| 100% | | 84.5 | 77.3 | 82.5 |
| 30% | IMP | **83.2** | 75.9 | **78.5** |
| | IMP+SPUR | 82.8 | **76.0** | 77.8 |
| | (GAP) | -0.4 | +0.1 | -0.7 |
| 10% | IMP | 77.8 | 72.7 | 68.7 |
| | IMP+SPUR | **79.3** | **73.3** | **70.6** |
| | (GAP) | +1.4 | +0.5 | +1.9 |
| 6% | IMP | 73.8 | 69.5 | 62.9 |
| | IMP+SPUR | **78.2** | **72.4** | **69.4** |
| | (GAP) | +4.4 | +2.9 | +6.5 |
| 3% | IMP | 68.9 | 64.2 | 58.0 |
| | IMP+SPUR | **75.6** | **70.1** | **66.3** |
| | (GAP) | +6.7 | +6.0 | +8.3 |

TABLE 2
QA PERFORMANCE COMPARISON BY SPUR

| Encoder Remaining Weights(%) | Method | Question Answering (EM/F1) | | |
|---|---|---|---|---|
| | | English (SQuAD) | Korean (KorQuAD) | French (FQuAD) |
| 100% | | 80.4/88.1 | 52.8/80.3 | 78.1/88.1 |
| 30% | IMP | **77.3/86.3** | 14.3/58.4 | 43.6/68.2 |
| | IMP+SPUR | 77.0/85.9 | **14.3/58.6** | **44.4/68.7** |
| | (GAP) | -0.4/-0.3 | -0.0/+0.2 | +0.8/+0.5 |
| 10% | IMP | 68.7/78.5 | 12.6/54.6 | 17.1/36.3 |
| | IMP+SPUR | **70.2/80.2** | **12.7/55.6** | **18.8/39.0** |
| | (GAP) | +1.5/+1.7 | +0.1/+1.0 | +1.8/+2.6 |
| 6% | IMP | 58.1/70.7 | 10.7/50.2 | 8.2/23.9 |
| | IMP+SPUR | **65.3/76.6** | **12.2/53.8** | **12.0/28.5** |
| | (GAP) | +7.1/+5.9 | +1.5/+3.6 | +3.8/+4.6 |
| 3% | IMP | 40.1/54.5 | 7.9/39.7 | 2.9/10.9 |
| | IMP+SPUR | **59.2/71.4** | **11.2/51.1** | **6.8/21.2** |
| | (GAP) | +19.2/+16.9 | +3.3/+11.4 | +3.9/+10.4 |

and IMP was observed to widen as the sparsity increases.

In the case of English NLI (MNLI) with a 10% sparsity setting, IMP yielded an accuracy of 77.8 and SPUR 79.3, showing an improvement of 1.4 points. As mentioned above, this performance gap widened as the sparisties increased. At 3% sparsity, IMP yielded 68.9, while adding SPUR yielded 75.6, widening the performance gap to 6.7 points. For Korean NLI (KorNLI) at the 3% setting, IMP yielded 64.2 with SPUR yielding 70.1 (improvement of 6.0 points); for French NLI (XNLI-FR) at the same rate, IMP scored 58.0, while adding SPUR yielded 66.3, for a performance gain of 8.3 points in Table 1.

Similar performance gains were also observed for Question Answering tasks. For English QA (SQuAD) at 10% sparsity, the F1 score for IMP was 78.5 and SPUR was 80.2 for a gain of 1.7 points. At the 3% setting, IMP was 54.3 and adding SPUR yielded 71.4 for a significant gain of 16.9 points in Table 2. Similar results followed for the Korean and French QA tasks.

As mentioned previously, since SPUR requires setting a final $\lambda$ value, we carried out the experiments with values of 10 and 100 and selected the best performing configuration. With the empirical observation that $\lambda$ value of 10 in the 30% sparsity setting performed the best regardless of the language and that a value of 100 performed best otherwise, we configured the $\lambda$ values accordingly.

### 4.4 Additional Experiments

The deviance function $D(W)$ used in the above experiments applies an $\ell_2$ penalty on the deviance statistic standardized by expected value. Additionally, $D_{L1S}(W)$ applies an $\ell_1$ penalty instead, $D_{L1}(W)$ applies no penalty at all, and $D_{L2}(W)$ applies an $\ell_2$ penalty without standardizing the deviance statistic.

Table 3 compares the performance of each deviance function setting on the English NLI [7] task for sparsities at 10% and 3%. Among them, SPUR was found to be the most effective compared to the other three deviance settings at each level of pruning.

In order to examine whether the performance differs according to the range in which regularization is applied, Table 4 shows the result of experimenting with applying it to just subsets of the model instead of all 6 dense layers of each BERT encoder layer. $BERT_{Q+K}$ only applies regularization on the query ($Q$) and key ($K$) matrices of the self-attention module, $BERT_Q$ only on the $Q$ matrix, and $BERT_K$ on the $K$ matrix. We found that the best performance resulted in utilizing the full set of dense layers.

### 4.5 Analysis

As in the example presented in RPP [2], visualizations of the second, third, and eleventh layers of the $Q$ and $K$ matrices by shading the removed weights in black are shown in Fig. 4 and Fig. 5. From these visualizations, one can again observe that the surviving weights corresponding to SPUR on the left form a grid structure. On the other hand, the surviving weights corresponding to IMP on the right are much more irregularly scattered. As the regularization term induces the weights to share similar values among the others on the same rows and columns, we can see that a grid structure forms as the more important rows and columns survive as entire groups.

TABLE 3
PERFORMANCE COMPARISON BY DEVIANCE D(W)

| Encoder Remaining Weights(%) | D(W) | MNLI (Accuracy) |
|---|---|---|
| 10% | **SPUR** | **79.3** |
| | L1S | 77.8 |
| | L1 | 77.7 |
| | L2 | 78.0 |
| 3% | **SPUR** | **75.6** |
| | L1S | 72.9 |
| | L1 | 74.5 |
| | L2 | 71.5 |

TABLE 4
PERFORMANCE COMPARISON BY REGULARIZATION DOMAIN

| Encoder Remaining Weights(%) | SPUR Domain | MNLI (Accuracy) |
|---|---|---|
| 10% | **BERT** | **79.3** |
| | $BERT_{Q+K}$ | 78.4 |
| | $BERT_Q$ | 77.9 |
| | $BERT_K$ | 78.0 |
| 3% | **BERT** | **75.6** |
| | $BERT_{Q+K}$ | 71.7 |
| | $BERT_Q$ | 70.4 |
| | $BERT_K$ | 70.2 |

This phenomenon is not without its potential limitations, however, since there is a potential risk that the weights as important as those on the grids can result in being ignored due to their particular positions. Because this has a potential to cause performance degradations, we performed an analysis comparing the variance in the values of the surviving weights in order to evaluate whether using SPUR was leading the weights to be distributed more uniformly.

Table 5 shows the representative statistics for these two cases. We first calculated the standard deviation and mean of the surviving weights for each weight matrix and computed a simple average of them. It may seem as though SPUR does induce a more uniform distribution of surviving weights, seeing that the average standard deviation was 0.008 with SPUR and 0.019 without SPUR. However, one should also consider that SPUR simply leads to smaller values overall, seeing that the average values of weights was 0.05 with SPUR and 0.108 without. We additionally considered the coefficient of variation (CV), a statistic often used for comparing variance of between distributions containing values of varying magnitudes from one another. The analysis showed that CV was 15.3% with SPUR and 17.9% without. With such a small difference, no significant evidence was found that SPUR causes the weight distributions to be distributed more uniformly.

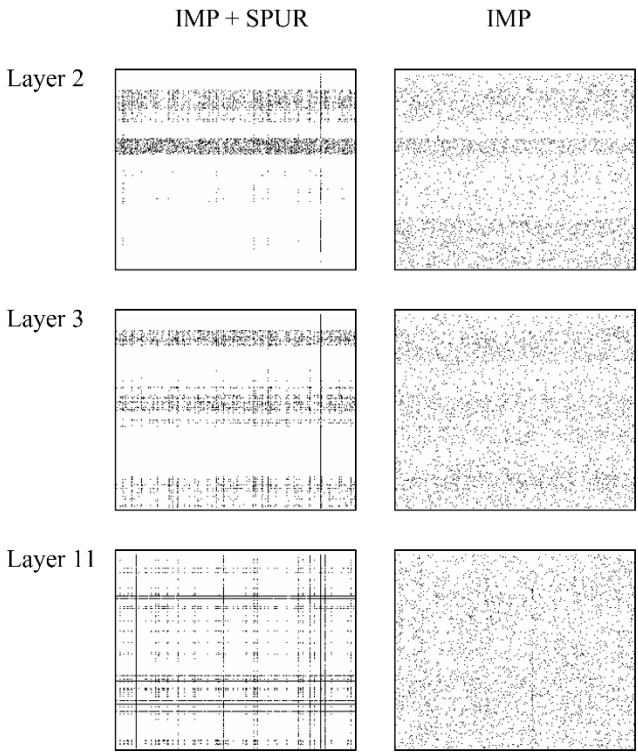

Fig. 4. Visualizations of the query matrices in the second, third, and eleventh layers of the encoder that was trained on the Korean NLI dataset then pruned to 5% sparsity. (surviving weights in black)

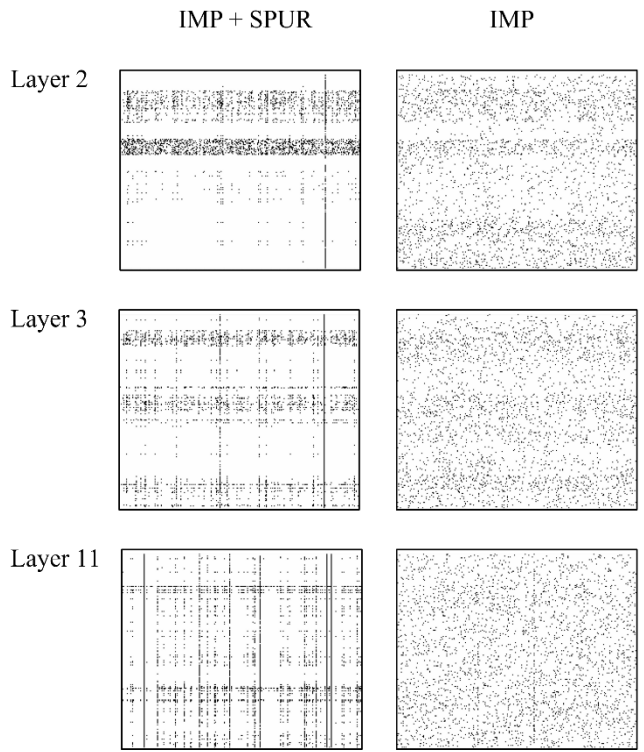

Fig. 5. Visualizations of the key matrices in the second, third, and eleventh layers of the encoder that was trained on the Korean NLI dataset then pruned to 5% sparsity. (surviving weights in black)

TABLE 5
STATISTICS OF SURVIVING WEIGHTS BY PRUNING METHOD

|   | IMP+SPUR | IMP |
|---|---|---|
| Avg. | 0.056 | 0.108 |
| Std. | 0.008 | 0.019 |
| CV | 15.3% | 17.9% |

TABLE 6
PERFORMANCE COMPARISON WITH OTHER APPROACHES

| Encoder Remaining Weights(%) | Method | English Tasks | |
|---|---|---|---|
| | | MNLI (Accuracy) | SQuAD (EM/F1) |
| 100% | | 84.5 | 80.4/88.1 |
| 30% | MvP | 80.7 | 72.5/82.4 |
| | IMP+SPUR | **82.8** | **77.0/85.9** |
| | (GAP) | +2.1 | +4.5/+3.5 |
| 10% | MvP | 79.3 | **71.9/81.7** |
| | IMP+SPUR | 79.3 | 70.2/80.2 |
| | (GAP) | - | -1.7/-1.5 |
| 3% | MvP | **76.1** | **65.2/76.2** |
| | IMP+SPUR | 75.6 | 59.2/71.4 |
| | (GAP) | -0.5 | -6.0/-4.8 |

With the Movement Pruning (MvP) [4] as a baseline that has been shown to perform very well, Table 6 shows the result of comparing it against SPUR in the English NLI (MNLI) and Question Answering tasks. SPUR performed better at 30% sparsity, while Movement Pruning performed better at 3% sparsity. It is worth noting again that the Movement Pruning required significantly more resource overhead since it requires a separate optimization of parameters at the same size as the actual model optimization.

## 5 CONCLUSION

Though extremely powerful, transformer-based models have notoriously large parameters; this is why the research community has been very actively studying methods that can effectively prune these models while retaining performance as much as possible. Among these methods, IMP has prevailed as one of the most widely-used methods, a process by which a model is pruned by iteratively removing the smallest weights and retraining the model. Visualizations of the inner weights in models that have been pruned by IMP have shown that the surviving weights tend to cluster in a grid structure, where most of them are found in a select few rows and columns. Though further research was necessary in exploring the use of these structured patterns in pruning models, there has not yet been such a study.

Recently, while the Movement Pruning has been developed to show very high performance, its limitations include the fact that it is immensely resource-intensive and that no research has shown that its performance generalizes to multilingual NLP tasks.

This study adds upon the IMP by adding a regularization term, resulting in a sizable performance increase and a unique propensity to preemptively induce the structured patterns as a result of pruning.

This study contributes the following. Firstly, we empirically show that the SPUR can enhance the performance of IMP, regardless of the language or the task at hand. Secondly, our study is the first to not only use regularization terms to preemptively induce the structured patterns, but also show that they are an effective attribute to model pruning methods. Thirdly, we've proposed a method that can achieve these performance gains from previous methods without the resource-intensive need for optimizing large score matrices in parallel.

One limitation of our study is that, despite how much hyperparameters can affect the resulting performance, we were not able to experiment with all possible combinations. Furthermore, though we have empirically shown that we can achieve greater retention of model performance by preemptively inducing structured patterns, further research is required in laying the theoretical foundations that explain this effect.

## REFERENCES

[1] S. Han, J. Pool, J. Tran, and W. Dally, "Learning both Weights and Connections for Efficient Neural Network," *Advances in Neural Information Processing Systems*, vol. 28, 2015.

[2] F.-M. Guo, S. Liu, F. S. Mungall, X. Lin, and Y. Wang, "Reweighted proximal pruning for large-scale language representation," *arXiv preprint arXiv:1909.12486*, 2019.

[3] P. Ganesh et al., "Compressing large-scale transformer-based models: A case study on bert," *arXiv preprint arXiv:2002.11985v1*, 2020.

[4] V. Sanh, T. Wolf, and A. Rush, "Movement Pruning: Adaptive Sparsity by Fine-Tuning," in *Advances in Neural Information Processing Systems*, 2020, vol. 33, pp. 20378–20389. [Online]. Available: https://proceedings.neurips.cc/paper/2020/file/eae15aabaa768ae4a5993a8a4f4fa6e4-Paper.pdf

[5] A. Vaswani et al., "Attention is All You Need," in *Proceedings of the 31st International Conference on Neural Information Processing Systems*, Red Hook, NY, USA, 2017, pp. 6000–6010.

[6] J. Devlin, M.-W. Chang, K. Lee, and K. Toutanova, "BERT: Pre-training of Deep Bidirectional Transformers for Language Understanding," in *Proceedings of the 2019 Conference of the North American Chapter of the Association for Computational Linguistics: Human Language Technologies, Volume 1 (Long and Short Papers)*, Minneapolis, Minnesota, Jun. 2019, pp. 4171–4186. doi: 10.18653/v1/N19-1423.

[7] A. Williams, N. Nangia, and S. Bowman, "A Broad-Coverage Challenge Corpus for Sentence Understanding through Inference," in *Proceedings of the 2018 Conference of the North American Chapter of the Association for Computational Linguistics: Human Language Technologies, Volume 1 (Long*

*Papers)*, New Orleans, Louisiana, Jun. 2018, pp. 1112–1122. doi: 10.18653/v1/N18-1101.

[8] A. Conneau *et al.*, "XNLI: Evaluating Cross-lingual Sentence Representations," in *Proceedings of the 2018 Conference on Empirical Methods in Natural Language Processing*, Brussels, Belgium, Oct. 2018, pp. 2475–2485. doi: 10.18653/v1/D18-1269.

[9] J. Ham, Y. J. Choe, K. Park, I. Choi, and H. Soh, "KorNLI and KorSTS: New Benchmark Datasets for Korean Natural Language Understanding," in *Findings of the Association for Computational Linguistics: EMNLP 2020*, Online, Nov. 2020, pp. 422–430. doi: 10.18653/v1/2020.findings-emnlp.39.

[10] P. Rajpurkar, J. Zhang, K. Lopyrev, and P. Liang, "SQuAD: 100,000+ Questions for Machine Comprehension of Text," in *Proceedings of the 2016 Conference on Empirical Methods in Natural Language Processing*, Austin, Texas, Nov. 2016, pp. 2383–2392. doi: 10.18653/v1/D16-1264.

[11] M. d'Hoffschmidt, W. Belblidia, Q. Heinrich, T. Brendlé, and M. Vidal, "FQuAD: French Question Answering Dataset," in *Findings of the Association for Computational Linguistics: EMNLP 2020*, Online, Nov. 2020, pp. 1193–1208. doi: 10.18653/v1/2020.findings-emnlp.107.

[12] S. Lim, M. Kim, and J. Lee, "KorQuAD1.0: Korean QA dataset for machine reading comprehension," *arXiv preprint arXiv:1909.07005*, 2019.

[13] M. Gordon, K. Duh, and N. Andrews, "Compressing BERT: Studying the Effects of Weight Pruning on Transfer Learning," in *Proceedings of the 5th Workshop on Representation Learning for NLP*, Online, Jul. 2020, pp. 143–155. doi: 10.18653/v1/2020.repl4nlp-1.18.

[14] E. J. Candes, M. B. Wakin, and S. P. Boyd, "Enhancing sparsity by reweighted ℓ 1 minimization," *Journal of Fourier analysis and applications*, vol. 14, no. 5, pp. 877–905, 2008.

[15] M. Zhu and S. Gupta, "To prune, or not to prune: exploring the efficacy of pruning for model compression," *arXiv preprint arXiv:1710.01878*, 2017.

[16] SKT Brain, "KoBERT," *GitHub repository*. GitHub, 2019. [Online]. Available: https://github.com/SKTBrain/KoBERT

[17] L. Martin *et al.*, "CamemBERT: a Tasty French Language Model," in *Proceedings of the 58th Annual Meeting of the Association for Computational Linguistics*, Online, Jul. 2020, pp. 7203–7219. doi: 10.18653/v1/2020.acl-main.645.